\documentclass[sigconf, nonacm]{acmart}

\usepackage{microtype}
\usepackage{graphicx}
\usepackage{subfigure}
\usepackage{booktabs}
\usepackage{hyperref}
\usepackage{url}
\usepackage{epsfig}
\usepackage{multirow}
\usepackage{enumitem}
\usepackage{balance}

\AtBeginDocument{%
  \providecommand\BibTeX{{%
    \normalfont B\kern-0.5em{\scshape i\kern-0.25em b}\kern-0.8em\TeX}}
}

\copyrightyear{2020}
\acmYear{2020}
\setcopyright{acmcopyright}\acmConference[CIKM '20]{Proceedings of the 29th ACM International Conference on Information and Knowledge Management}{October 19--23, 2020}{Virtual Event, Ireland}
\acmBooktitle{Proceedings of the 29th ACM International Conference on Information and Knowledge Management (CIKM '20), October 19--23, 2020, Virtual Event, Ireland}
\acmPrice{15.00}
\acmDOI{10.1145/3340531.3411875}
\acmISBN{978-1-4503-6859-9/20/10}

\begin{document}
\fancyhead{}

\title{NagE: Non-Abelian Group Embedding for Knowledge Graphs}

\author{Tong~Yang}\authornote{Equal Contribution} 
\affiliation{%
	\institution{Department of Physics, \\ Boston College,
	Chestnut Hill, \\ Massachusetts, USA}
}
\author{Long~Sha}\authornotemark[1]
\affiliation{%
	\institution{Department of Computer Science, Brandeis University,
	Waltham, Massachusetts, USA}
}
\author{Pengyu~Hong}
\affiliation{%
	\institution{Department of Computer Science, Brandeis University,
	Waltham, Massachusetts, USA}
}

\begin{abstract}

We demonstrated the existence of a group algebraic structure hidden in relational knowledge embedding problems, which suggests that a group-based embedding framework is essential for designing embedding models. Our theoretical analysis explores merely the intrinsic property of the embedding problem itself hence is model independent. Motivated by the theoretical analysis, we have proposed a group theory-based knowledge graph embedding framework, in which relations are embedded as group elements, and entities are represented by vectors in group action spaces. We provide a generic recipe to construct embedding models associated with two instantiating examples: \textbf{SO3E} and \textbf{SU2E}, both of which apply a continuous non-Abelian group as the relation embedding. Empirical experiments using these two exampling models have shown state-of-the-art results on benchmark datasets.

\end{abstract}

\ccsdesc[500]{Computing methodologies~Knowledge representation and reasoning}
\ccsdesc[300]{Computing methodologies~Statistical relational learning}

\maketitle

\section{Introduction}


Knowledge graphs (KGs) are prominent structured knowledge bases for many downstream semantic tasks~\cite{hao-2017}. A KG contains an entity set $\mathcal{E} = \{e_i\}$, which correspond to vertices in the graph, and a relation set $\mathcal{R} = \{r_k\}$, which forms edges. The entity and relation sets form a collection of factual triplets, each of which has the form $(\mathbf{e}_i, \mathbf{r}_k, \mathbf{e}_j)$ where $\mathbf{r}_k$ is the relation between the head entity $\mathbf{e}_i$ and the tail entity $\mathbf{e}_j$.  Since large scale KGs are usually incomplete due to missing links (relations) amongst entities, an increasing amount of recent works~\cite{transe, dismult, complex, transr} have devoted to the graph completion (i.e., link prediction) problem by exploring a low-dimensional representation of entities and relations. 

More formally, each relation $\mathbf{r}$ acts as a mapping $\mathbf{O}_{\mathbf{r}}[\cdot]$ from its head entity $\mathbf{e}_1$ to its tail entity $\mathbf{e}_2$:
\begin{align}
    \mathbf{r}: \mathbf{e}_1 \mapsto
    \mathbf{O}_{\mathbf{r}}[\mathbf{e}_1]=:\mathbf{e}_2.
\end{align}
The original KG dataset represents these mappings in a tabular form, and the task of knowledge graph embedding (KGE) is to find a better representation for these abstract mappings. For example, in the TransE model~\cite{transe}, relations and entities are embedded in the same vector space, and the operation $\mathbf{O}_{\mathbf{r}}[\cdot]$ is simply a vector summation: $\mathbf{O}_{\mathbf{r}}[\mathbf{e}]=\mathbf{e}+\mathbf{r}$. In general, the operation could be either linear or nonlinear, either pre-defined or learned.
Importantly, graph completion relies on the fact that relations are not independent. For example, the hypernym and hyponym are inverse to each other; while kinship relations usually support mutual inferences. These dependencies would impose constraints on the operation design. Previous studies~\cite{rotate, dihedral} have concerned some specific cases of inter-relation dependencies, including (anti-)symmetry and compositional relations. 

In this work, however, we attempt to deliver a high-level analysis from a general perspective. More particularly, we ask the following three questions:
\begin{enumerate}
    \item What constraints/requirements does a general KGE task impose on embeddings?
    \item What kind of embedding method would satisfy these constraints?
    \item How to explicitly construct embedding models?
\end{enumerate}
Note that the first question concerns general KGE tasks rather than specific datasets nor embedding models, and, therefore, requires an analysis including all possible knowledge graph structures. We find that, to accommodate all possible KG datasets, there are five requirements for the relation embedding: \textbf{\textit{closure, identity, inverse, associativity,}} and \textbf{\textit{non-commutativity}}. The first four coincide with the algebraic definition of groups in mathematics, and imply a direct answer to the second question: embedding all relations into a group manifold (and designing mapping operations as group actions) would automatically satisfy all requirements; 
in addition, the last requirement, non-commutativity, further suggests implementing non-Abelian groups for the most general KGE tasks.
The third question asks for a general recipe to embed relations as group elements.

One main contribution of this work is it provides a framework for addressing the KG embedding problem from a novel and more rigorous perspective: the group-theoretic perspective. We prove that the intrinsic structure of general KGE tasks coincides with the complete definition of groups. To our best knowledge, this is the first proof that rigorously legitimates the application of group theory in KG embedding. With this framework, we also establish connections with many existing models (see Sec \ref{3.3}), including: TransE~\cite{transe}, TransR~\cite{transr}, TorusE~\cite{toruse}, RotatE~\cite{rotate}, ComplEx~\cite{complex}, DisMult~\cite{dismult}. 

The remaining sections are organized as following: 
Section~\ref{relatedworks} mentions some related works, and emphasizes the distinction between our analysis and others; 
in Section~\ref{general_kge}, we answer the first two question proposed above, achieving a conclusion that continuous non-Abelian groups suit general KGE tasks well, which leads to the exampling continuous non-Abelian group embedding models (\textbf{NagE}) in later sections; 
in Section~\ref{implementing}, we provide a general recipe for group-embedding approach of KGE problems, associated with two novel instantiating models: \textbf{SO3E} and \textbf{SU2E}, which completes the answer for all three questions; 
we demonstrate the power of the proposed embedding framework by comparing the experimental results of our two example models with other state-of-the-art models in Section~\ref{experiments}, and conclude all discussions in Section~\ref{conclusion}.

\section{Related Works}\label{relatedworks}

From the group theory perspective, our work may be related to the TorusE~\cite{toruse}, the RotatE~\cite{rotate}, DihEdral~\cite{dihedral}, and QuatE~\cite{quate} models. 

The TorusE model frames the KG entity embedding in a Lie group manifold to deal with the \emph{compactness problem}. The authors proved that the additive nature of the operation in the TransE model contradicts the entity regularization. However, if a non-compact embedding space is used, entities must be regularized to prevent the divergence of negative scores. Therefore the TorusE model used $n$-torus, a compact manifold, as the entity embedding space. In other words, TorusE embeds entities in group manifolds, while our work embeds relations as group manifolds. 

In the DihEdral model, the $D_{4}$ group the author used plays the same role as groups in our work: group elements correspond to relations. The motivation of DihEdral is to resolve the non-Abelian composition (i.e., the compositional relation formed by $\mathbf{r}_1$ and $\mathbf{r}_2$ would change if the two are switched). Nevertheless, DihEdral applies a discrete group $D_8$ for relation embedding while using a continuous entity embedding space, which may suffer two problems as discussed in the later Section \ref{3.3}. The RotatE model was designed to accommodate symmetric, inversion, and (abelian) composition of triplets at the same time. 
Different from these previous works, our work does not target at one or few specific cases but aims at answering the more general question: finding the generative principle of all possible cases, and thus to provide guidance for model designs that can accommodate all cases. 

More importantly, in most preceding works related to groups~\cite{dihedral, cai2019group}, group theory serves as an alternative perspective to explain the efficiency of specific models; while the theoretical analysis in our current work in Section~\ref{general_kge} is model/dataset independent and is merely initiated by KGE tasks themselves, and the group embedding approach automatically emerges as the natural method which could satisfy all constraints for a general KGE task.
To our the best of our knowledge, this is the first proof that rigorously legitimates the implementation of group theory in KG embeddings.

Another interesting connection refers to QuatE ~\cite{quate} model, where the authors proposed quaternions (also octonions and sedenions in the appendix), which served as an extension of complex numbers for KGE. The intuition of their model was not related to group theory at all. However, in Model Analysis (Section 5.3 in [4]), the authors stated that: “Normalizing the relation to unit quaternion is a critical step for the embedding performance.” While this was empirically observed, an explicit reason is absent. This phenomena can be easily understood from the group theory perspective, as long as one realizes the mathematical correspondence: SU(2) group is an isomorphism to unit quaternions. This explains the necessity of applying “normalization” on quaternions: only unit quaternions are consistent with the group structure (SU(2) in specific), while non-unit ones cannot form a group. One of our newly proposed model, \textbf{SU2E}, is therefore closely related to QuatE model with "unit-scaling", although our proposal does not concern different number systems at all. It is also worth to mention that when comparing with other proceeding models, the authors applied a completely different criteria for the QuatE experiments: a "type constraint" was introduced in their experiments, which filtered out a significant portion of challenging relation types during the evaluation phase. As a contrast, our \textbf{SU2E} model proposed in later sections investigated the similar setting but compared with other models under the common criteria (without "type constraint"), and showed superior results as a \emph{continuous non-Abelian group embedding} for the first time.

\section{Group theory in relational embeddings}\label{general_kge}
In this section, we formulate the group-theoretic analysis for relational embedding problems. 
Firstly, as most embedding models represent objects (including entities and relations) as vectors, the task of operation design thus can be stated as finding proper transformations of vectors.
Secondly, as we mentioned in the introduction, our ultimate goal of reproducing the whole knowledge graph using atomic triplets further requires certain types of local patterns to be accommodated. We now discuss these structures, which in the end naturally leads to the definition of groups in mathematics.

\subsection{Hyper-relation Patterns: relation-of-relations}\label{hyper_relation}

One difficulty of generating the whole knowledge graph from atomic triplets lies in the fact that different relations are not independent of each other. The task of relation inference relies exactly on their mutual dependency. In other words, there exist certain \emph{relation of relations} in the graph structure, which we term as \emph{hyper-relation patterns}. A proper relation embedding method and the associated operations should be able to capture these hyper-relations.

Now instead of studying exampling cases one by one, we ask the most general question: what are the most fundamental hyper-relations? The answer is quite simple and only contains two types, namely, \emph{inversion} and \emph{composition}:
\begin{itemize}
    \item \textbf{Inversion}: given a relation $\mathbf{r}$, there \emph{may exist} an inversion $\mathbf{\bar{r}}$, such that, $\forall \mathbf{e}_1, \mathbf{e}_2\in \mathcal{E}$:
    \begin{align}
        \mathbf{r}:\mathbf{e}_1\mapsto\mathbf{e}_2
        \;\; \longrightarrow \;\; 
        \mathbf{\bar{r}}:\mathbf{e}_2\mapsto\mathbf{e}_1.
    \end{align}
    The inversion captures any relation path with a length equal to 1 (in the unit of relations).
    
    \item \textbf{Composition}: given two relations $\mathbf{r}_1$ and $\mathbf{r}_2$, there \emph{may exist} a third relation $\mathbf{r}_3$, such that, $\forall \mathbf{e}_1, \mathbf{e}_2, \mathbf{e}_3\in \mathcal{E}$:
    \begin{align}\label{composition}
    \left\{
        \centering
        \begin{tabular}{l}
             $\mathbf{r}_1:\mathbf{e}_1\mapsto\mathbf{e}_2$\\ $\mathbf{r}_2:\mathbf{e}_2\mapsto\mathbf{e}_3$
        \end{tabular}
    \right. 
    \quad \longrightarrow \quad 
    \mathbf{r}_3:\mathbf{e}_1\mapsto\mathbf{e}_3.
    \end{align}
    Any relation paths longer than 1 can be captured by a sequence of compositions.
\end{itemize}
One may notice the phrase \emph{may exist} in the above definition, this simply emphasizes that the existence of these derived conceptual relations $\bar{\mathbf{r}}$ and $\mathbf{r}_3$ depends on the \textbf{specific KG dataset}; while, on the other hand, to accommodate general KG datasets, the \textbf{embedding space} should always contains the mathematical representations of these conceptual relations.

An important feature of KG is that with the above two hyper-relations, one could generate any local graph pattern and eventually the whole graph, as relational paths with arbitrary length have been captured.  
Note the term of \emph{inversion} and \emph{composition} might have different meanings from ones in other works: most existing works study triplets to analyze hyper relations, while the definition we provide above is based purely on relations. This is more general in the sense that any conclusion derived would not depend on entities at all, and some different hyper relations could, therefore, be summarized as a single one. For example, there are enormous discussions on \emph{symmetric} triplets and \emph{anti-symmetric} triplets~\cite{rotate}, which are defined as:
\begin{align*}
    \centering
    \left.
    \begin{tabular}{rccr}
        \textbf{symmetric}:  &  $(\mathbf{e}_1, \mathbf{r}, \mathbf{e}_2)$ & $\longrightarrow$ & $(\mathbf{e}_2, \mathbf{r}, \mathbf{e}_1)$,\\
        \textbf{anti-symmetric}:  &  $(\mathbf{e}_1, \mathbf{r}, \mathbf{e}_2)$ & $\longrightarrow$ & $\neg(\mathbf{e}_2, \mathbf{r}, \mathbf{e}_1)$.\\ 
    \end{tabular}
    \right.
\end{align*}
In fact, if for any choice of $\mathbf{e}_{1,2}$, one could produce a symmetric pair of true triplets using $\mathbf{r}$, this would imply a property of $\mathbf{r}$ itself, and in which case, one could then simply derive:
\begin{align}
    \mathbf{\bar{r}} = \mathbf{r}.
\end{align}
This is a special case of the inversion hyper-relation; and similarly, the anti-symmetric case simply implies $\mathbf{\bar{r}} \neq \mathbf{r}$, which is quite common, and does not require extra design. 
The deep reason for discussing hyper-relations which relies merely on relations rather than triplets is that the logic of relation inference problem itself is not entity-dependent.

\subsection{Emergent Group Theory}\label{theory}
To accommodate both general inversions and general compositions, we now derive explicit requirements on the relation embedding model. We start by defining the \emph{product} of two relations $\mathbf{r}_1$ and $\mathbf{r}_2$: $\mathbf{r}_1\cdot\mathbf{r}_2$, as subsequently "finding the tail" twice according to the two relations, i.e. 
\begin{align}\label{product}
\mathbf{O}_{\mathbf{r}_1\cdot\mathbf{r}_2}\big[\cdot\big] :=\mathbf{O}_{\mathbf{r}_1}\big[\mathbf{O}_{\mathbf{r}_2}[\cdot]\big].
\end{align}
With the above definition, \eqref{composition} can be rewritten as: $\mathbf{r}_3 = \mathbf{r}_1 \cdot \mathbf{r}_2$.
One would realize that the following properties should be supported by a proper embedding model:
\begin{enumerate}
    \item \textbf{Inverse element}: to allow the possible existence of inversion, the elements $\mathbf{\bar{r}}$ should also be an element living in the same relation-embedding space\footnote{Given a graph, not all inversions correspond to meaningful relations, but an embedding model should be able to capture this possibility in general.}.
    
    \item \textbf{Closure}: to allow the possible existence of composition, in general, the elements $\mathbf{r}_1 \cdot \mathbf{r}_2$ should also be an element living in the same relation-embedding space\footnote{Given a graph, not all compositions correspond to meaningful relations, but an embedding model should be able to capture this possibility in general.}.
    
    \item \textbf{Identity element}: the possibly existing inversion and composition together define another special and unique relation:
    \begin{align}
        \mathbf{i} = \mathbf{r} \cdot \mathbf{\bar{r}}, \qquad \forall \mathbf{r}\in\mathcal{R}.
    \end{align}
    This element should map any entity to itself, and thus we call it \emph{identity element}.
    
    \item \textbf{Associativity}: In a relational path with the length longer than three (containing three or more relations $\{r_1, r_2, r_3, ...\}$), as long as the sequential order does not change, the following two compositions should produce the same result:
    \begin{align}\label{associativity}
        (\mathbf{r}_1 \cdot \mathbf{r}_2)\cdot \mathbf{r}_3
        = \mathbf{r}_1 \cdot (\mathbf{r}_2\cdot \mathbf{r}_3).
    \end{align}
    The associativity is actually rooted in our definition of $\mathbf{r}_1\cdot\mathbf{r}_2$ in \eqref{product} through the subsequent operating sequence in the entity space,
    from which, we can derive directly that:
    \begin{align}
        \mathbf{O}_{(\mathbf{r}_1 \cdot \mathbf{r}_2)\cdot \mathbf{r}_3}\big[\cdot\big] &= \mathbf{O}_{\mathbf{r}_1\cdot\mathbf{r}_2}\big[ \mathbf{O}_{\mathbf{r}_3}[\cdot]\big]  \nonumber \\
        &= \mathbf{O}_{\mathbf{r}_1}\big[ \mathbf{O}_{\mathbf{r}_2}\big[ \mathbf{O}_{\mathbf{r}_3}[\cdot]\big]\big]\\
        &= \mathbf{O}_{\mathbf{r}_1}\big[ \mathbf{O}_{\mathbf{r}_2\cdot\mathbf{r}_3}[\cdot]\big]
        = \mathbf{O}_{\mathbf{r}_1 \cdot (\mathbf{r}_2\cdot \mathbf{r}_3)}\big[\cdot\big], \nonumber
    \end{align}
    which then leads to the association \eqref{associativity}.
    To help readers understand the practical meaning of associativity in real life cases, here we provide a simple example of the relational associativity: 
    \begin{align*}
        \mathbf{r}_1 = \textbf{\text{isBrotherOf}}, \;  
        \mathbf{r}_2 = \textbf{\text{isMotherOf}}, \;  
        \mathbf{r}_3 = \textbf{\text{isFatherOf}}.
    \end{align*}
    Meanwhile, the following compositions are also meaningful:
    \begin{align*}
        \mathbf{r}_1\cdot\mathbf{r}_2 = \textbf{\text{isUncleOf}}, \quad
        \mathbf{r}_2\cdot\mathbf{r}_3 = \textbf{\text{isGrandmotherOf}}.
    \end{align*}
    In this example, one could easily see that:
    \begin{align*}
        (\mathbf{r}_1 \cdot \mathbf{r}_2)\cdot \mathbf{r}_3
        = \mathbf{r}_1 \cdot (\mathbf{r}_2\cdot \mathbf{r}_3)
        = \textbf{\text{isGranduncleOf}}.
    \end{align*}
    This is a simple demonstration of the associativity.

    \item \textbf{Commutativity/Nonconmmutativity}: In general, commuting two relations in a composition, i.e. $\mathbf{r}_1\cdot\mathbf{r}_2 \leftrightarrow \mathbf{r}_2\cdot\mathbf{r}_1$, may compose either the same or different results.
    We provide a simple illustrative examples for non-commutative compositions. Consider the following real world kinship:
    \begin{align}
        \mathbf{r}_1 = \textit{\text{isMotherOf}}, \quad  
        \mathbf{r}_2 = \textit{\text{isFatherOf}}.
    \end{align}
    Clearly, the composition $\mathbf{r}_1\cdot\mathbf{r}_2$ and $\mathbf{r}_2\cdot\mathbf{r}_1$ correspond to $\textit{\text{isGrandmotherOf}}$ and $\textit{\text{isGrandfatherOf}}$ relations respectively, which are different. This is a simple example of non- commutative cases.
    In real graphs, any cases may exist, and a proper embedding method should be able to accommodate both.
\end{enumerate}
The first four properties are exactly the definition of a \textbf{\textit{group}}. In other words, \emph{the group theory automatically emerges from the relational embedding problem itself, rather than being applied manually.} This is a quite convincing evidence that group theory is indeed the most natural language for relational embeddings if one aims at ultimately reproducing all possible local patterns in graphs.
Besides, the fifth property on \emph{commutativity/nonconmmutativity} are actually termed as \textbf{abelian/non-Abelian} in the group theory language. Since abelian is only a special case, to accommodate all possibilities, one should, in general, consider a \textbf{non-Abelian group for the relation embedding}, and guarantee at the same time it contains at least one nontrivial abelian subgroup. We would term the corresponding embedding method as \textbf{NagE: the non-Abelian group embedding method}.

More explicitly, given a graph, to implement a group structure in embedding, one should embed all relations as group elements, which are parametrized by certain group parameters. For instance: 
the translation group $T$ can be parametrized by a real number $\delta$. And correspondingly, due to its vector nature, the embedding of entities could be regarded as a \textbf{representation (rep)} space of the same group. For the translation group, $\mathbb{R}$ (the real field) is a rep space of $T$.

This suggests the group representation theory is useful in knowledge graph embedding problems when talking about entity embeddings, and we leave this as a separate topic for subsequent works later. In the later section, we provide a general recipe for the graph embedding implementation.

\subsection{Embedding models using different groups}\label{3.3}
In this section, we discuss embedding methods using different groups, from simple ones as $T$ (the translation group) and $U(1)$, to complicated ones including $SU(2)$, $GL(n, \mathbb{V})$ (where $\mathbb{V}$ could be any type of fields), or even \textit{Aff}$(\mathbb{V})$. It is important to note that, in practice, \textbf{continuous groups} are more reasonable than discrete ones, due to the two following reasons:
\begin{itemize}
    \item The entity embedding space is usually continuous, which matches reps of the continuous group better. If used to accommodate a discrete group, a continuous space always contains infinite copies of irreducible reps of that group, which makes the analysis much more difficult.
    
    \item When training the embedding models, a gradient-based optimization search would be applied in the parameter space. However, different from continuous groups whose group parameter are also continuous, the parametrization of a discrete group uses discrete values, which brings in extra challenges for the training procedure.
\end{itemize}
With the two reasons above, we thus mainly consider continuous groups which are more reasonable choices. The other important feature of a group is commutativity, which we would mention for each group below. 
Besides the relational embedding group $\mathcal{G}$, the entity embedding space and the similarity measure also need to be determined. As discussed above, the entity embedding should be a proper rep space of $\mathcal{G}$. 
While for similarity measure $d(\cdot)$, we choose among the popular ones including $L_p$-norms ($L_p$) and the $\cos$-similarity ($\cos$), and a complete score function $s_{\mathbf{r}}(\mathbf{e_1}, \mathbf{e_2})$ for a triplet $(\mathbf{e_1}, \mathbf{r}, \mathbf{e_2})$ would be the distance from the relation-transformed head entity to the tail entity. 
One would notice many choices reproduce precedent works, and we show two examples below.

\subsubsection{Example group: \texorpdfstring{$T$}{Lg}}\label{T}
One could use $n$-copies of $T$, the translation group, for the relation embedding. This is a \textit{noncompact abelian} group. The simplest rep-space would be the real field $\mathbb{R}$, which should also appear $n$ times as $\mathbb{R}^n$. The group embedding then produces the following embedding vectors:
\begin{align*}
    \mathbf{e} \quad \Longrightarrow \quad 
    \Vec{v}_{\mathbf{e}}&= \big(x_1, x_2, \;\;\cdots, \;\;x_n\big), \qquad \forall \mathbf{e}\in \mathcal{E}; \nonumber \\
    \mathbf{r} \quad \Longrightarrow \quad 
    \Vec{v}_{\mathbf{r}} &= \big(\delta_1, \; \delta_2, \;\;\,\cdots, \;\;\delta_n\big), \qquad \forall \mathbf{r}\in \mathcal{R};
\end{align*}
both of which are $n$-dim. Here both $x_i$ and $\delta_i$ are real numbers. In a triplet, the relation $\Vec{v}_r$ acts as an addition vector added to the head entity $\mathbf{e}_1$.
If one further chooses $L_p$-norm as the similarity measure, the complete score function $s_{\mathbf{r}}(\mathbf{e_1}, \mathbf{e_2})$ would be:
\begin{align}
    \|(\Vec{v}_{\mathbf{r}} + \Vec{v}_{\mathbf{e_1}}) - \Vec{v}_{\mathbf{e_2}}\|_p,
\end{align}
this actually corresponds to the well-known \textbf{TransE} model~\cite{transe}.
There was a regularization in the original TransE model that changes the entity rep-space, which however has been removed in many later works by properly bounding the negative scores.

\subsubsection{Example group: \texorpdfstring{$U(1)$}{Lg}}\label{u1}

One could use $n$-copies of $U(1)$, the 1-dim unitary transformation group, for the relational embedding. This is a \textit{compact abelian} group. 
The simplest rep-space would be the real field $\mathbb{C}$, which should also appear $n$ times as $\mathbb{C}^n$. The group embedding then produces the following embedding vectors:
\begin{align*}
    \mathbf{e} \quad \Longrightarrow \quad 
    \Vec{v}_{\mathbf{e}}&= \big(x_1, x_2, \;\;\cdots, \;\;x_n\big), \qquad \forall \mathbf{e}\in \mathcal{E}; \nonumber \\
    \mathbf{r} \quad \Longrightarrow \quad 
    \Vec{v}_{\mathbf{r}} &= \big(\phi_1,  \phi_2, \;\;\,\cdots, \;\phi_n\big), \qquad \forall \mathbf{r}\in \mathcal{R};
\end{align*}
where $x_i$ is a complex number containing a both real and imaginary part, while $\phi_i$ is a phase variable take values from $0$ to $2\pi$. Therefore the entity-embedding dimension is $2n$, while the relation dimension is $n$.
In a triplet, the relation $\Vec{v}_{\mathbf{r}}$ acts as a phase shift on the head entity $\mathbf{e}_1$. In a matrix form, one could define $\mathbf{R}_{\mathbf{r}}$ as the diagonal matrix with the $i$-th diagonal element being $e^{i\phi_i}$.
If one further chooses $L_p$-norm as the similarity measure, the complete score function $s_{\mathbf{r}}(\mathbf{e_1}, \mathbf{e_2})$ would be:
\begin{align}
    \|\mathbf{R}_{\mathbf{r}}\cdot\Vec{v}_{\mathbf{e}_1} - \Vec{v}_{\mathbf{e}_2}\|_p =
    \|\big[e^{i\Vec{v}_{\mathbf{r}}}\big]\circ\Vec{v}_{\mathbf{e}_1} - \Vec{v}_{\mathbf{e}_2}\|_p,
\end{align}
where $\circ$ means a Hadamard product.
This precisely leads to the \textbf{RotatE} model~\cite{rotate}. 

On the other hand, one could also use the $n$-torus $\mathbb{T}^n$ as the rep-space:
\begin{align*}
    \mathbf{e} \quad \Longrightarrow \quad 
    \Vec{v}_{\mathbf{e}}&= \big(\theta_1, \theta_2, \;\;\cdots, \;\;\theta_n\big), \qquad \forall \mathbf{e}\in \mathcal{E}; \nonumber \\
    \mathbf{r} \quad \Longrightarrow \quad 
    \Vec{v}_{\mathbf{r}} &= \big(\phi_1,  \phi_2, \;\;\,\cdots, \,\phi_n\big), \qquad \forall \mathbf{r}\in \mathcal{R};
\end{align*}
where $\theta_i$ represents a coordinate on the torus. Still using the $L_p$-norm similarity measure, the complete score function $s_{\mathbf{r}}(\mathbf{e_1}, \mathbf{e_2})$ now is:
\begin{align}
    \|\mathbf{R}_{\mathbf{r}}\cdot\Vec{v}_{\mathbf{e}_1} - \Vec{v}_{\mathbf{e}_2}\|_p =
    \|e^{i\Vec{v}_{\mathbf{r}}}\circ e^{i\Vec{v}_{\mathbf{e}_1}} - e^{i\Vec{v}_{\mathbf{e}_2}}\|_p,
\end{align}
which leads to the \textbf{TorusE} model~\cite{toruse}.
In the original implementation of TorusE, there is an additional projection $\pi$ from $\mathbb{R}^n$ to $\mathbb{T}^n$.\footnote{Due to the special relation between $T$ and $U(1)$, i.e. $U(1)\cong T/(2\pi\mathbb{Z})$, one could also regard TorusE as an implementation of group-embedding with $T$, which is more similar to the motivation in the original paper~\cite{toruse}.}

\subsubsection{A summary of some example groups}
We summarize the results of several chosen examples in Table~\ref{tab:groups}.

\begin{table*}
\centering
\begin{tabular}{|c|c|c|c|c|c|}
\hline
Group & Space & Abelian & $d(\cdot)$ & Studied & Related Work \\
\hline
$T$ & $\mathbb{R}^n$ & YES & $L_p$ & $\checkmark$ & TransE~\cite{transe}\\
\hline
$U(1)$  & $\mathbb{C}^n$ & YES & $L_p$ & $\checkmark$ & RotatE~\cite{rotate}\\
\hline
$U(1)$  & $\mathbb{T}^n$ & YES & $L_p$ & $\checkmark$ & TorusE~\cite{toruse}\\
\hline
$SO(3)$  & $\mathbb{R}^{3 n}$ & NO & $L_p$ & -- & --\\
\hline
$SU(2)$  & $\mathbb{C}^{2 n}$ & NO & $L_p$ & -- & -- \\
\hline
$GL(1, \mathbb{R})$  & $\mathbb{R}^n$ & YES & $\cos$ & $\checkmark$ & DisMult~\cite{dismult}\\
\hline
$GL(1, \mathbb{C})$  & $\mathbb{C}^n$ & YES & $\cos$ & $\checkmark$ & ComplEx~\cite{complex}\\
\hline
$GL(n, \mathbb{R})$  & $\mathbb{R}^n$ & NO & $\cos$ & -- & RESCAL~\cite{rescal}\\
\hline
\textit{Aff}$(\mathbb{R}^n)$ & $\mathbb{R}^n$ & NO & $L_p$ & -- & TransR~\cite{transr}\\
\hline
$D_4$  & $\mathbb{R}^n$ & NO & $L_p$ & $\checkmark$ & DihEdral~\cite{dihedral}\\
\hline
\end{tabular}
\caption{Examples of the group embedding.}
\label{tab:groups}
\end{table*}

Note in the Table~\ref{tab:groups}, some groups have not been studied, but there are still some existing models which use a quite similar embedding space; while the major gap, between the existing models and their group embedding counterparts, is the constraint of group structures on the parametrization.
For example, implementing group embedding with $GL(n, \mathbb{R})$, the $n$-dim general linear groups defined on field-$\mathbb{R}$, would lead to a model similar to RESCAL~\cite{rescal}. However, the original RESCAL model does not have a built-in group structure: it uses arbitrary $n\times n$ real matrices, some of which may not be invertible, and hence are not group elements in $GL(n,\mathbb{R})$. It is, therefore, worth to add the extra invertible constraint in RESCAL, which requests matrices constructed through group parametrization rather than assigned arbitrary matrix elements. A similar analysis holds for the affine group Aff$(\mathbb{R}^n)$.

\section{Group Embedding for Knowledge Graphs}\label{implementing}
In this section, we would firstly provide a general recipe for the group embedding implementation, and then provide two explicit examples of \textbf{NagE}, both of which apply a continuous non-Abelian group that has not been studied in any precedent works before.

\subsection{A general group embedding recipe}\label{recipe}
We summarize the group embedding procedure as following:
\begin{enumerate}
    \item Given a graph, choose a proper group $\mathcal{G}$ for embedding. The choice may concern property of the task, such as commutativity and so on. And as stated above, in most general cases, a non-Abelian continuous group should be proper.
    \item Choose a rep-space for the entity embedding. For simplicity, one could use multiple ($n$) copies of the same rep $\rho$, which is the case of most existing works. Suppose $\rho$ is a $p$-dim rep, then the total dimension of entity embedding would be $pn$, which is written as a vector $\Vec{v}_{\mathbf{e}}$. Roughly speaking, $k$ captures the relational structure and $n$ encodes other feature.
    \item Choose a proper parametrization of $\mathcal{G}$, that is, choose a set of parameters indexing all group elements in $\mathcal{G}$. Suppose the number of parameters required to specify a group element is $q$, then the total dimension of relation embedding $\Vec{v}_{\mathbf{r}}$ would be $qn$. A group element can now be expressed as a block-diagonal matrix $\mathbf{R}_{\mathbf{r}}$, with each block $\mathbf{M}_i$ being a $p\times p$ matrix whose entries are determined by the vector $\Vec{v}_{\mathbf{r}}$.
    \item Choose a similarity measure $d(\cdot)$, the score value $s_{\mathbf{r}}(\mathbf{e}_1,\mathbf{e}_2)$ of a triplet $(\mathbf{e}_1,\mathbf{r},\mathbf{e}_2)$ is then:
    \begin{align}
        s_{\mathbf{r}}(\mathbf{e}_1,\mathbf{e}_2) \equiv d\big( \mathbf{R}_{\mathbf{r}}\cdot\Vec{v}_{\mathbf{e}_1}, \;\; \Vec{v}_{\mathbf{e}_2} \big)
    \end{align}
\end{enumerate}
Below we demonstrate the group embedding approach by implementing it with exampling continuous non-Abelian groups. As shown in table~\ref{tab:groups}, two simple continuous non-Abelian groups that have not been studied are $SO(3)$ and $SU(2)$, we will implement them as relation embedding manifolds, which, as a result, produce two \textbf{NagE models}.

\subsection{\textbf{SO3E}: NagE with group \texorpdfstring{$SO(3)$}{Lg}}
The 3D special orthogonal group $SO(3)$ is one of the simplest continuous non-Abelian group. 
As an illustrative demonstration, we construct an embedding model with $SO(3)$ structure and implement it in real experiments. Following the general recipe above, after determining the group $\mathcal{G}=SO(3)$, we choose a proper rep-space for entity embedding: $[\mathbb{R}^3]^{\otimes n}$, which consists $n$-copies of $\mathbb{R}^3$. Each $\mathbb{R}^3$ subspace transforms as the standard rep-space of $SO(3)$. All relations thus act as $3n\times 3n$ block diagonal matrix, with each block being a $3\times 3$ complex matrix carrying the standard representation of $SO(3)$.

Next, we choose a proper parametrization of $SO(3)$. Instead of the more general angular momentum parametrization, due to our choice of using the standard representation, we could parameterize the $SO(3)$ elements using Euler angles $(\phi, \theta, \psi)$, which is easier for implementation.

Put all together, our group embedding is then fixed as:
\begin{align*}
    \mathbf{e} \; \Longrightarrow \; 
    \Vec{v}_{\mathbf{e}}&= \big(x_1, y_1, z_1, \;\;\cdots, \;\;x_n, y_n, z_n\big), \quad \forall \mathbf{e}\in \mathcal{E}; \nonumber \\
    \mathbf{r} \; \Longrightarrow \; 
    \Vec{v}_{\mathbf{r}} &= \big(\phi_1, \theta_1, \psi_1, \;\; \cdots, \;\phi_n, \theta_n, \psi_n\big), \quad \forall \mathbf{r}\in \mathcal{R};
\end{align*}
both of which are $3n$-dim. In a triplet, the relation vector $\Vec{v}_{\mathbf{r}}$ acts as a block diagonal matrix $\mathbf{R}_{\mathbf{r}}$, with each block matrix $M_i$ acting in the subspace of $(x_i, y_i, z_i)$:
\begin{align}\label{large_matrix}
    \setlength{\tabcolsep}{4pt}
    \centering
    \left[
    \begin{tabular}{ccc|ccc|ccc|ccc}
     {} & {} & {} & {} & {} & {} & {} & {} & {} & {} & {} & {} \\
     {} & $\mathbf{M}_1$ & {} & {} & $0$ & {} & {} & $\ldots$ & {} & {} & $0$ & {} \\
     {} & {} & {} & {} & {} & {} & {} & {} & {} & {} & {} & {} \\
     \hline
     {} & {} & {} & {} & {} & {} & {} & {} & {} & {} & {} & {} \\
     {} & $0$ & {} & {} & $\mathbf{M}_2$ & {} & {} & $\ldots$ & {} & {} & $0$ & {} \\
     {} & {} & {} & {} & {} & {} & {} & {} & {} & {} & {} & {} \\
     \hline
     {} & {} & {} & {} & {} & {} & {} & {} & {} & {} & {} & {} \\
     {} & $\vdots$ & {} & {} & $\vdots$ & {} & {} & $\ddots$ & {} & {} & $\vdots$ & {} \\
     {} & {} & {} & {} & {} & {} & {} & {} & {} & {} & {} & {} \\
     \hline
     {} & {} & {} & {} & {} & {} & {} & {} & {} & {} & {} & {} \\
     {} & $0$ & {} & {} & $0$ & {} & {} & $\ldots$ & {} & {} & $\mathbf{M}_n$ & {} \\
     {} & {} & {} & {} & {} & {} & {} & {} & {} & {} & {} & {} 
    \end{tabular}
    \right]
    \left[
    \begin{tabular}{c}
         $x_1$ \\
         $y_1$ \\
         $z_1$ \\
         \hline
         $x_2$ \\
         $y_2$ \\
         $z_2$ \\
         \hline
         \\
         \vdots \\
         \\
         \hline
         $x_n$ \\
         $y_n$ \\
         $z_n$
    \end{tabular}
    \right].
\end{align}
And each $3\times 3$ block $\mathbf{M}_i$ is parametrized as following:
\begin{align}
    \mathbf{M}_i^{11} &= \quad\cos{\psi_i}\cos{\phi_i} - \cos{\theta_i}\sin{\psi_i}\sin{\phi_i}\nonumber\\ 
    \mathbf{M}_i^{12} &= \quad\cos{\psi_i}\sin{\phi_i} + \cos{\theta_i}\cos{\psi_i}\cos{\phi_i}\nonumber\\ 
    \mathbf{M}_i^{13} &= \quad\sin{\psi_i}\sin{\theta_i}\nonumber\\ 
    \mathbf{M}_i^{21} &=-\sin{\psi_i}\cos{\phi_i} - \cos{\theta_i}\sin{\psi_i}\cos{\phi_i}\nonumber\\ 
    \mathbf{M}_i^{22} &=-\sin{\psi_i}\sin{\phi_i} + \cos{\theta_i}\cos{\psi_i}\cos{\phi_i}\nonumber\\ 
    \mathbf{M}_i^{23} &= \quad\cos{\psi_i}\sin{\theta_i}\nonumber\\ 
    \mathbf{M}_i^{31} &= \quad\cos{\psi_i}\sin{\theta_i} \nonumber\\ 
    \mathbf{M}_i^{32} &=-\cos{\psi_i}\cos{\theta_i}\nonumber\\ 
    \mathbf{M}_i^{33} &= \quad\cos{\theta_i}
\end{align}

\subsection{\textbf{SU2E}: NagE with group \texorpdfstring{$SU(2)$}{Lg}}
The 2D special unitary group $SU(2)$ is another simple continuous non-Abelian group. 
We choose the rep-space for entity embedding as: $[\mathbb{C}^2]^{\otimes n}$, which consists $n$-copies of $\mathbb{C}^2$. Each $\mathbb{C}^2$ subspace transforms as the standard rep-space of $SU(2)$. All relations thus act as $2n\times 2n$ block diagonal matrix, with each block being a $2\times 2$ complex matrix carrying the standard representation of $SU(2)$.

Next, we choose a proper parametrization of $SU(2)$. An analysis with the corresponding Lie algebra $\mathfrak{su}(2)$ shows that any group element could be written as~\cite{lie}:
\begin{align}
    e^{i\alpha [\hat{\mathbf{n}}\cdot\Vec{\mathbf{J}}]} = 
    \cos{\alpha}\hat{1} + i\sin{\alpha}\hat{\mathbf{n}}\cdot\Vec{\mathbf{J}},
\end{align}
where $\alpha$ is a rotation angle taken from $[0,2\pi]$, and $\hat{\mathbf{n}}$ is a unit vector on $S^2$, represented by two other angles $(\theta, \phi)$; moreover, the symbol $\hat{1}$ means an identity matrix, and $\Vec{\mathbf{J}}$ are three generators of the group: $(\mathbf{J}_x, \mathbf{J}_y, \mathbf{J}_z)$, which, in the standard rep have the following form:
\begin{align*}
    \centering
    J_x = 
    \left[
    \begin{tabular}{c c}
    0 & 1 \\
    1 & 0
    \end{tabular}
    \right],\; 
    J_y = 
    \left[
    \begin{tabular}{c c}
    0 & -i \\
    i & 0
    \end{tabular}
    \right],\; 
    J_z = 
    \left[
    \begin{tabular}{c c}
    1 & 0 \\
    0 & -1
    \end{tabular}
    \right].
\end{align*}
Put all together, our group embedding is then fixed as:
\begin{align*}
    \mathbf{e} \; \Longrightarrow \; 
    \Vec{v}_{\mathbf{e}}&= \big(x_1, y_1, \; x_2, y_2, \;\cdots, \;x_n, y_n\big), \quad \forall \mathbf{e}\in \mathcal{E}; \nonumber \\
    \mathbf{r} \; \Longrightarrow \; 
    \Vec{v}_{\mathbf{r}} &= \big(\alpha_1, \hat{\mathbf{n}}_1, \alpha_2, \hat{\mathbf{n}}_2,\; \cdots, \alpha_n, \hat{\mathbf{n}}_n\big), \quad \forall \mathbf{r}\in \mathcal{R};
\end{align*}
where $x_i$ and $y_i$ are complex numbers, and $\alpha_i$ and $\hat{\mathbf{n}}_i=(\theta_i, \phi_i)$ represent angles. 
In a triplet, the relation $\Vec{v}_{\mathbf{r}}$ acts as a block diagonal matrix $\mathbf{R}_{\mathbf{r}}$, with each block matrix $M_i$ acting in the subspace of $(x_i, y_i)$. And each $2\times 2$ block $\mathbf{M}_i$ is parametrized as~\cite{lie}:
\begin{align*}
    \centering
    \left[
    \begin{tabular}{c c}
         $\cos{\alpha_i} + i\sin{\alpha_i}\sin{\theta_i}$ & 
         $ie^{-i\phi_i}\cdot\sin{\alpha_i}\cos{\theta_i}$
         \\
         $ie^{-i\phi_i}\cdot\sin{\alpha_i}\cos{\theta_i}$ & 
         $\cos{\alpha_i} - i\sin{\alpha_i}\sin{\theta_i}$
    \end{tabular}
    \right]. 
\end{align*}

\subsection{Similarity measure and loss function}
We choose $L_2$-norm as the similarity measure $d(\cdot)$ to compute the score value:
\begin{align}
     s_{\mathbf{r}}(\mathbf{e}_1,\mathbf{e}_2) = \|\mathbf{R}_{\mathbf{r}}\cdot\Vec{v}_{\mathbf{e}_1} - \Vec{v}_{\mathbf{e}_2}\|_2
\end{align}
We design the model loss function for a triple $(\mathbf{e_1}, \mathbf{r}, \mathbf{e_2})$ as follows:
\begin{align*}
    &L =-\log{\sigma\left[\gamma-s_{\mathbf{r}}(\mathbf{e_1}, \mathbf{e_2})\right]}\nonumber\\
    &\qquad -\sum_{i=1}^{n} p\left(\mathbf{e_{1i}^{\prime}}, \mathbf{r}, \mathbf{e_{2i}^{\prime}}\right) \log{\sigma\left[s_r\left(\mathbf{e_{1i}}^{\prime}, \mathbf{e_{2i}}^{\prime}\right)-\gamma\right]} \\
    &p\left(\mathbf{e_{1j}^{\prime}}, \mathbf{r}, \mathbf{e_{2j}^{\prime}} |\left\{\left(\mathbf{e_{1i}}, \mathbf{r}, \mathbf{e_{2i}}\right)\right\}\right) =\frac{e^{\alpha[\gamma-s_{\mathbf{r}}(\mathbf{e'_{1j}}, \mathbf{e'_{2j}})]}}{\sum_{i} e^{\alpha [\gamma-s_{\mathbf{r}}(\mathbf{e'_{1i}}, \mathbf{e'_{2i}})]}}
\end{align*}
where $\sigma$ is the Sigmoid function, $\gamma$ is the margin used to prevent over-fitting. $\mathbf{e_{1i}^{\prime}}$ and $\mathbf{e_{2i}^{\prime}}$ are negative samples while $p\left(\mathbf{e_{1i}^{\prime}}, r, \mathbf{e_{2i}^{\prime}}\right)$ is the adversarial sampling mechanism with temperature $\alpha$ we adopt self-adversarial negative sampling setting from ~\cite{rotate}. We term the resulting model as \textbf{SO3E} and \textbf{SU2E} respectively for the above two groups.
We mention other implementation details in the next section.

\section{Experiments}\label{experiments}

\subsection{Experimental Setup}

\paragraph{Datasets: }

\begin{table*}[!ht]
\begin{center}
\begin{tabular}{|c|c|cccc|cccc|c|}
\hline
\multirow{2}*{Group} & \multirow{2}*{Commutativity} & \multicolumn{4}{|c|}{WN18}  & \multicolumn{4}{|c|}{FB15k} & \multirow{2}*{Example} \\ \cline{3-10}

& & MRR & H@1 & H@3 & H@10 & MRR & H@1 & H@3 & H@10 & \\ \hline

T & Abelian & 0.495 & 0.113 & 0.888 & 0.943 & 0.463 & 0.297 & 0.578 & 0.749 & TransE \\
U$(1)$ & Abelian & \underline{0.949} & \textbf{0.944} & 0.952 & \underline{0.959} & \textbf{0.797} & \textbf{0.746} & \underline{0.830} & \underline{0.884} & RotatE \\ 
U$(1)$ & Abelian & 0.947 & \underline{0.943} & 0.950 & 0.954 & 0.733 & 0.674 & 0.771 & 0.832 & TorusE \\  
GL$(1,\mathbb{R})$ & Abelian & 0.822 & 0.728 & 0.914 & 0.936 & 0.654 & 0.546 & 0.733 & 0.824 & DistMult \\ 
GL$(1,\mathbb{C})$ & Abelian & 0.946 & 0.942 & 0.949 & 0.954 & 0.692 & 0.599 & 0.759 & 0.840 & ComplEx \\ \hline
SO$(3)$ & non-Abelian & \textbf{0.950} & \textbf{0.944} & \underline{0.953} & \textbf{0.960} & \underline{0.794} & \underline{0.740} & \textbf{0.831} & \textbf{0.886} & SO3E \\ 
SU$(2)$ & non-Abelian & \textbf{0.950} & \textbf{0.944} & \textbf{0.954} & \textbf{0.960} & 0.791 & 0.734 & \textbf{0.831} & \textbf{0.886} & SU2E \\ \hline
\end{tabular}
\caption{\label{tab:1815}Link prediction on WN18 and FB15k (\textbf{bold} represent the best scores, \underline{underlined} represent the second best).}
\end{center}
\end{table*}

\begin{table*}[!ht]
\begin{center}
\begin{tabular}{|c|c|cccc|cccc|c|}
\hline
\multirow{2}*{Group} & \multirow{2}*{Commutativity} & \multicolumn{4}{|c|}{WN18RR}  & \multicolumn{4}{|c|}{FB15k-237} & \multirow{2}*{Example} \\ \cline{3-10}

& & MRR & H@1 & H@3 & H@10 & MRR & H@1 & H@3 & H@10 & \\ \hline

T & Abelian & 0.226 & - & - & 0.501 & 0.294 & - & - & 0.465 & TransE \\
U$(1)$ & Abelian & \underline{0.476} & 0.428 & \underline{0.492} & 0.571 & \underline{0.338} & 0.241 & 0.375 & \textbf{0.533} & RotatE \\ 
GL$(1,\mathbb{R})$ & Abelian & 0.430 & 0.390 & 0.440 & 0.490 & 0.241 & 0.155 & 0.263 & 0.419 & DistMult \\ 
GL$(1,\mathbb{C})$ & Abelian & 0.440 & 0.410 & 0.460 & 0.510 & 0.247 & 0.158 & 0.275 & 0.428 & ComplEx \\ \hline
SO$(3)$ & non-Abelian & \textbf{0.477} & \textbf{0.432} & \textbf{0.493} & \underline{0.574} & \textbf{0.340} & \textbf{0.244} & \textbf{0.378} & 0.530 & SO3E \\ 
SU$(2)$ & non-Abelian & \underline{0.476} & \underline{0.429} & \textbf{0.493} & \textbf{0.575} & \textbf{0.340} & \underline{0.243} & \underline{0.376} & \underline{0.532} & SU2E \\ \hline
\end{tabular}
\caption{\label{tab:237rr}Link prediction on WN18RR and FB15k-237 (\textbf{bold} represent the best scores, \underline{underlined} represent the second best).}
\end{center}
\end{table*}

The most popular public knowledge graph datasets include FB15K~\cite{freebase} and  WN18~\cite{wordnet}. FB15K-237~\cite{fb237} and WN18RR~\cite{wnrr} datasets were derived from these two, in which the inverse relations were removed. FB15K dataset is a huge knowledge base with general facts containing 1.2 billion instances of more than 80 million entities. For benchmarking, usually, a frequency filter was applied to obtain occurrence larger than 100 resulting in 592,213 instances with 14,951 entities and 1,345 relation types. WN18 was extracted from WordNet~\cite{wordnet} dictionary and thesaurus, the entities are word senses and the relations are lexical relations between them. It has 151,442 instances with 40,943 entities and 18 relation types.
\paragraph{Evaluation Protocols: }
We use three categories of protocols for evaluations, namely, cut-off Hit ratio (H@N), Mean Rank(MR) and Mean Reciprocal Rank (MRR). H@N measures the ratio of correct entities predictions at a top $n$ prediction result cut-off. Following the baselines used in recent literature, we chose $n=1,3,10$. MR evaluates the average rank among all the correct entities. MRR is the average rank inverse rank of the correct entities.

\paragraph{Implementation Details: }
We implemented our models using pytorch\footnote{https://www.pytorch.org} framework and experimented on a server with an Nvidia Titan-1080 GPU. The Adam~\cite{Adam} optimizer was used with the default $\beta_{1}$ and $\beta_{2}$ settings. A learning rate scheduler observing validation loss decrease was used to reduce learning rate by half after patience of 3000. Batch-size was set at 1024.  We did a grid search on the following hyper-parameters: embedding dimension $d\in\{100, 250, 400, 500\}$; learning rate $\eta\in\{3e-4, 1e-4, 3e-5, 1e-5, 3e-6\}$; number of negative samples during training $n_{neg}\in\{128, 256, 512\}$; adversarial negative sampling temperature $\alpha \in\{0.5, 0.75, 1.0, 1.25\}$; loss function margin $\gamma \in \{6, 9, 12, 20, 22, 24, 26\}$.

\subsection{Results and Model analysis}

Empirical results on FB15k and WN18 are reported in Table \ref{tab:1815}. We compared the embedding results of different groups, including $T$, $U(1)$, $GL(1, \mathbb{R})$, $GL(1,\mathbb{C})$, $SO(3)$ and $SU(2)$, which are mainly categorized by the commutativity. As discussed in Sec.~\ref{3.3}, the former four groups have been implicitly applied in existing models. 
For $SO(3)$ and $SU(2)$, we report the result of our own experiments. Results of the other models are taken from their original literature: TransE using group $T$ was proposed in \cite{transe}; RotatE using group $U(1)$ was proposed in \cite{rotate} while TorusE with the same group was proposed in \cite{toruse}; group $GL(1,\mathbb{V})$ was implemented in DisMult~\cite{dismult} with $\mathbb{V}=\mathbb{R}$ and in ComplEx~\cite{complex} with $\mathbb{V}=\mathbb{C}$.

Results on datasets FB15K-237 and WN18RR are demonstrated in Table \ref{tab:237rr} respectively. We remove TorusE from the tables due to the absence of results in the original work, and refer to \cite{nguyen2017novel} for TransE.

In the FB15k dataset, the main hyper-relation is anti-/symmetry and inversion. The dataset has a vast amount of unique entities. Shown in Table \ref{tab:1815}, the RotatE model achieved good performance in this dataset. 
SO3E and SU2E achieved comparable result across the metrics. On the other hand, since inversion relations are removed in FB15k-237, the dominant portion of hyper-relations becomes the composition. We can see RotatE fail on this task due to non-Abelian hyper-relations. Shown in Table \ref{tab:237rr}, the continuous non-Abelian group method SO3E and SU2E outperformed most of the metrics.

In the WN18 dataset, SO3E and SU2E outperformed all the baselines on all metrics shown in Table \ref{tab:1815}. The WN18RR dataset removes the inversion relations from WN18, left only 11 relations and most of them are symmetry patterns. We can see from Table \ref{tab:237rr}, SO3E and SU2E model performed well due to their non-Abelian nature.

Drawn from the experiments, two factors significantly impact the embedding model performance: the embedding dimension, and group attributes (including commutativity and continuity). As theoretically analyzed in Section \ref{theory}, and empirically shown above, continuous non-Abelian groups are more reasonable choices for general tasks. It is important to note that SO3E and SU2E proposed above are exampling models for our group embedding framework, and they use the simplest continuous non-Abelian groups. Much more efforts could be devoted in this direction in the future.

\section{Conclusion and Future Work}\label{conclusion}

We proved for the first time the emergence of a group definition in the KG representation learning. This proof suggests that relational embeddings should respect the group structure. 
A novel theoretic framework based on group theory was therefore proposed, termed as the \emph{group embedding} of relational KGs. Embedding models designed based on our proposed framework would automatically accommodate all possible hyper-relations, which are building-blocks of the link prediction task. 

From the group-theoretic perspective, we categorize different embedding groups regarding commutativity and the continuity and empirically compared their performance. We also realize that many recent models correspond to embeddings using different groups. Generally speaking, a continuous non-Abelian group embedding should be powerful for a generic KG completion task. We demonstrate this idea by examining two simple exampling models: \textbf{SO3E} and \textbf{SU2E}. With $SO(3)$ and $SU(2)$ as embedding groups, our models showed promising performance in challenging tasks where hyper-relations become crucial.

In the proposed framework, beside embedding relations as group elements, entity embeddings live in different representation space of the corresponding group. And therefore an investigation of group representation theory in entity embedding is highly demanded. We leave this in future works.
On the other hand, although empirical evaluations focus on linear models, it is important to note that the proof of the group structure only relies on the KG task itself. This means our conclusion also works for more general models, including neural-network-based ones. Beyond KG embeddings, the same analysis could be applied to other representation learning where intrinsic relational structures are prominent. An implementation of group structures in more general cases would be very interesting.

\section*{Acknowledgement}
Funding for the shared GPU-computing facility used in this research was provided by NSF OAC 1920147.

\bibliographystyle{ACM-Reference-Format}
\balance
\bibliography{ref}

\end{document}